\pdfoutput=1

\documentclass{article} 
\usepackage{iclr2024_conference,times}

\usepackage{amsmath,amsfonts,bm}

\def\eqref#1{equation~\ref{#1}}

\def\1{\bm{1}}

\def\vb{{\bm{b}}}
\def\vc{{\bm{c}}}

\def\vx{{\bm{x}}}

\def\mA{{\bm{A}}}

\DeclareMathAlphabet{\mathsfit}{\encodingdefault}{\sfdefault}{m}{sl}
\SetMathAlphabet{\mathsfit}{bold}{\encodingdefault}{\sfdefault}{bx}{n}

\usepackage{hyperref}
\usepackage{url}
\usepackage{pgfplots}
\usepackage{enumitem}
\usepackage[normalem]{ulem}

\usepackage{tikz}
\usetikzlibrary{positioning}
\usepackage{listings}
\usepackage[capitalise]{cleveref}
\usepackage{subcaption}

\lstdefinestyle{mystyle}{
    basicstyle=\ttfamily\small,
    breaklines=true
}

\usepackage{xcolor}

\title{OptiMUS: Optimization Modeling Using MIP Solvers and Large Language Models}

\author{
Ali AhmadiTeshnizi, Wenzhi Gao, \& Madeleine Udell  \\
Department of Management Science and Engineering\\
Stanford University\\
Stanford, CA 94305, USA \\
\texttt{\{teshnizi,gwz,udell\}@stanford.edu} \\
\AND
}

\iclrfinalcopy

\pgfplotsset{compat=1.18}
\begin{document}

\maketitle

\begin{abstract}

Optimization problems are pervasive across various sectors, from manufacturing and distribution to healthcare. However, most such problems are still solved heuristically by hand rather than optimally by state-of-the-art solvers, as the expertise required to formulate and solve these problems limits the widespread adoption of optimization tools and techniques. We introduce OptiMUS, a Large Language Model (LLM)-based agent designed to formulate and solve MILP problems from their natural language descriptions. OptiMUS is capable of developing mathematical models, writing and debugging solver code, developing tests, and checking the validity of generated solutions. To benchmark our agent, we present NLP4LP, a novel dataset of linear programming (LP) and mixed integer linear programming (MILP) problems. Our experiments demonstrate that OptiMUS solves nearly twice as many problems as a basic LLM prompting strategy. OptiMUS code and NLP4LP dataset are available at 
\href{https://github.com/teshnizi/OptiMUS}{https://github.com/teshnizi/OptiMUS}
\end{abstract}

\section{Introduction}

Optimization problems are common in many fields like operations, economics, engineering, and computer science. 
Important applications of optimization include reducing energy use in smart grids, improving supply chains, or increasing profits in algorithmic trading \citep{OptModelingApp, antoniou2007practicalOptimization}. 
Major advances in optimization algorithms over the last several decades have led to reliable and efficient optimization
methods for a wide variety of structured optimization problems, including linear programming (LP) and mixed-integer linear programming (MILP) among many others. 
Unfortunately, optimization modeling --- transforming a business problem into a mathematical optimization problem in standard form ---
still requires expert knowledge.
This expertise gap prevents many organizations from using optimization, even when it could significantly improve their operations. E
xamples include inventory management in supermarkets, patient operations in hospitals, transportation policies in small municipalities, energy management in local solar farms, and operations in small businesses or NGOs \citep{saghafian2015operationsHospital, aastrup2010fortyRetail, yao2020optimizationTransport, shakoor2016wakeFarm}. 
Automating optimization modeling would allow sectors that can not afford access to optimization experts 
to improve efficiency using optimization techniques. 

\begin{figure}[t]
    \centering
    \includegraphics[width=\textwidth]{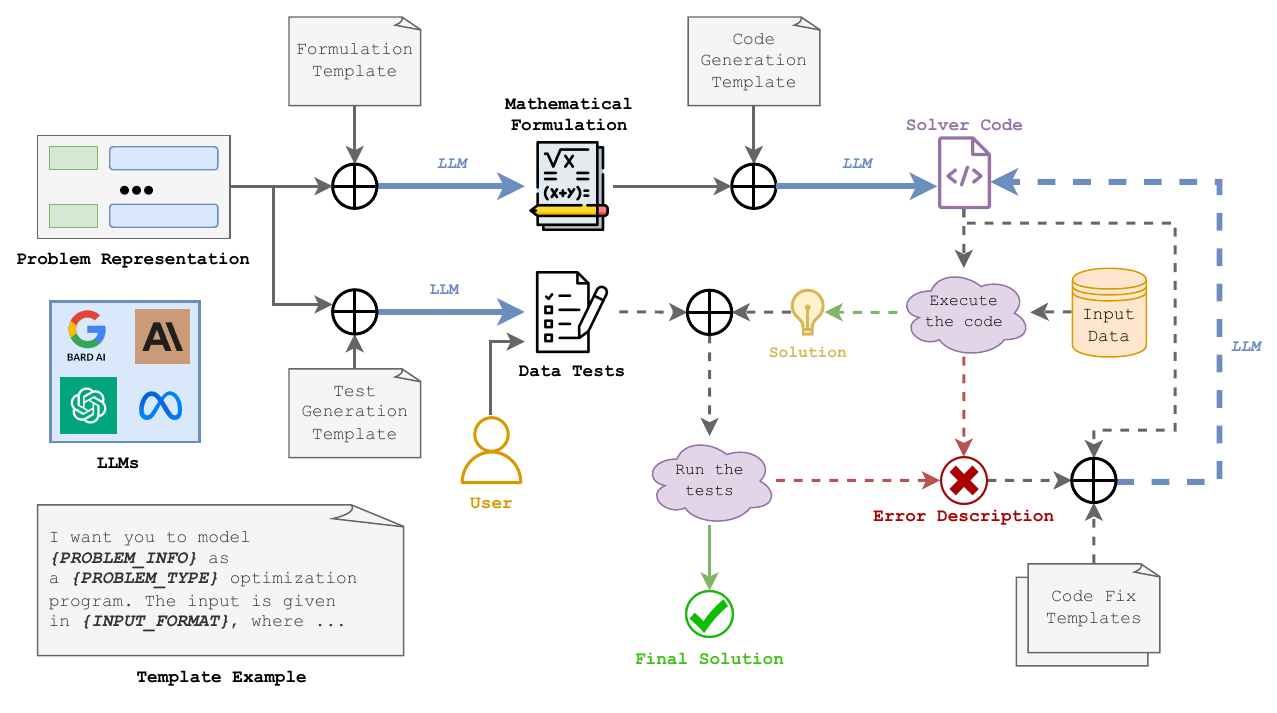}
    \caption{An illustration explaining how OptiMUS uses various components to effectively model and solve optimization problems. First, a mathematical formulation is generated from the problem representation. The solver code is then generated based on the formulation. The code is executed to generate and save a solution to file. The solution is then tested on a set of unit tests generated by the LLM and revised by the user. If the code does not run or fails the tests it is passed to the LLMs along for the relevant error code for revision until it is fixed (dashed lines might be executed multiple times). Otherwise, the output is selected as the final solution. An example template is shown in the bottom left corner.}
    \label{fig:agent_structure}
\end{figure}

Large language models (LLMs) offer a promising way to make optimization more accessible. LLMs have demonstrated the capability to understande, generate, and interpret natural language for many tasks. They make it easier to formulate problems and set up solutions, making expert knowledge more widely available. However, the role of LLMs in the optimization landscape is still unexplored, mainly due to their novelty and the absence of comprehensive benchmarks. To explore the capabilities and limitations of LLMs in optimization, this paper makes the following contributions:


\begin{itemize}[leftmargin=*]
    \item We introduce a novel dataset, NLP4LP, comprising human-expert formulations of 52 LP and MILP optimization problems, annotated with their solutions, code to check optimality, and sample formulations of the problem in markdown and in code. To construct this dataset, we introduce a standardized format to represent optimization problems in natural language.
    \item We present OptiMUS, an LLM-based agent to formulate and solve optimization problems. \cref{fig:agent_structure} demonstrates the structure of OptiMUS.
    \item We develop techniques to improve the quality of OptiMUS and demonstrate their effectiveness, including automated data augmentation via problem rephrasing and self-fixing of solutions via automated testing and debugging. Using these techniques, OptiMUS increases the solve rate by 91\% compared to direct prompting.
\end{itemize}

By integrating the capabilities of LLMs with optimization techniques, our work aims to democratize access to optimization across application domains, extending the reach and utility of optimization. 

This paper builds on recent progress in Large Language Models (LLMs) and optimization. 
A more comprehensive review of this work is deferred to \cref{related_work}, and we herein on ideas most closely related to our topic.
In a very recent paper, \cite{chen2023diagnosing} develop a chatbot to help users detect and fix infeasible optimization problems.
In contrast to our work, their agent takes a \texttt{Pyomo}  code rather than a natural language description as input, 
and acts as an AI assistant rather than as a solver.
\citep{yang2023LLMsAsOptimizers} use LLMs to directly generate solutions to optimization problems,
focusing on the problem of modifying prompts to improve performance.
In contrast to our work, their method does not rely on solvers, coding, or any other tools.
Their model can not address problems with medium or large input data sizes because 
1) even the context size of LLMs is very limited compared to the input data size of many optimization problems and 
2) LLMs' performance substantially decreases as the input context grows longer, even for explicitly long-context models \citep{liu2023lostinthemiddle}.

This paper is organized as follows:  \cref{challenges} discusses the challenges of using LLMs to solve optimization problems; \cref{methodology} describes the details of our LLM-based optimization agent; \cref{dataset} outlines the dataset creation process and statistics; \cref{experiments} presents the experiments and analysis; \cref{related_work} explores the related work; and \cref{conclusion} concludes the paper with future directions and implications. The appendix includes prompts, details on the experiments' setup, and further analysis.

\section{Challenges of Optimization Modeling using LLMs}
\label{challenges}
Optimization problems are defined mathematically by an objective function and a set of constraints. 
For example, an MILP can be written as

\[
\begin{array}{ll}
\mbox{minimize} & \vc^T \vx\\ 
\mbox{subject to} & \mA \vx \leq \vb\\
& x_i \in \mathbb{Z},~ i \in \mathcal{I}
\end{array}
\]
An optimization workflow consists of 1) formulating an optimization problem in standard form
by identifying its objective and constraints, and then 
2) solving the problem, generally using code that calls an optimization solver.
Formulation is often a challenging task even for experts in optimization.
Different formulations can lead to significantly different solving times and
enable the use of different solvers or solution techniques \citep{BoydConvex}.
One important skill for an optimization expert is to identify assumptions or relaxations that 
allow for casting the problem as a well-studied problem type, such as MILP, which
enables the use of well-developed optimization solvers.
Crafting such an efficient formulation often requires specialized knowledge \citep{ConicRelaxationSurvey, RelaxationOptimalFlow, roubivcek2020relaxation, SemidefiniteRelaxation, LagrangianRelaxationSurvey}. 

Given the formulation, an optimization expert must choose a solver. Each solver has its own interface and functionalities \citep{achterberg2019gurobi, diamond2016cvxpy, manual1987ibmcplex}. However, user manuals for these solvers are often hundreds of pages, making them hard to fully understand. 
This complexity often makes it challenging to choose the right solver and arrive at a high-quality implementation, much less to switch solvers in order to find the one best suited for a given problem.

Recent progress in Natural Language Processing (NLP) has led to the development of large language models (LLMs) useful in various tasks like answering questions, summarizing text, translating languages, and coding \citep{openai2023gpt4, touvron2023llama, chowdhery2022palm, wei2023chainofthought}. Connections to other software tools extends the reach and accuracy of LLMs, as demonstrated by plugins for code writing and execution \citep{paranjape2023art, wei2023chainofthought}. Consequently, leveraging LLMs for optimization modeling is a logical progression. A basic approach is to describe a problem to an LLM and prompt it to write the solver code. However, this direction faces several key challenges:

\begin{itemize}[leftmargin=*]

    \item \textbf{Variable Representations:} An optimization problem can be described in many ways. For example, a user might use different terms (vehicle vs. car vs. truck vs. carrier), notations (\textit{price} and \textit{capacity} vs. $p$ and $c$ vs. $x$ and $y$), or even omit assumptions based on common sense (capacity of a vehicle is nonnegative, number of employees is an integer, etc.). While humans can easily adapt to these variations, LLMs may struggle depending on their training data and methods.
    \item \textbf{Handling Large-Scale Problems:} LLMs have a limited context size. Even for long-context models, performance substantially decreases as the input context grows longer \citep{liu2023lostinthemiddle}. However, the specification of an optimization problem often involves a large amount of numerical data, such as attributes of customers or of goods sold, which could easily require gigabytes of memory. 
    
    \item \textbf{Unreliable Outputs:} The solutions provided by LLMs are not always reliable. The generated code may be incorrect or even not executable. It is especially challenging to verify the solution when the code runs but the output is incorrect. For instance, if the code runs and claims the problem is unbounded, perhaps a constraint has been accidentally omitted from the formulation? Moreover, writing tests to check the correctness of the output can also be complicated and time-consuming. 
    
\end{itemize}

Solving optimization problems with language models requires a multi-faceted approach combining advances in natural language processing, optimization theory, and software engineering. The following sections introduce our LLM-based optimization agent OptiMUS, demonstrate how OptiMUS mitigates these challenges, and present empirical evidence of its effectiveness and limitations.

\begin{figure}[t]
    \centering
    \includegraphics[width=\textwidth]{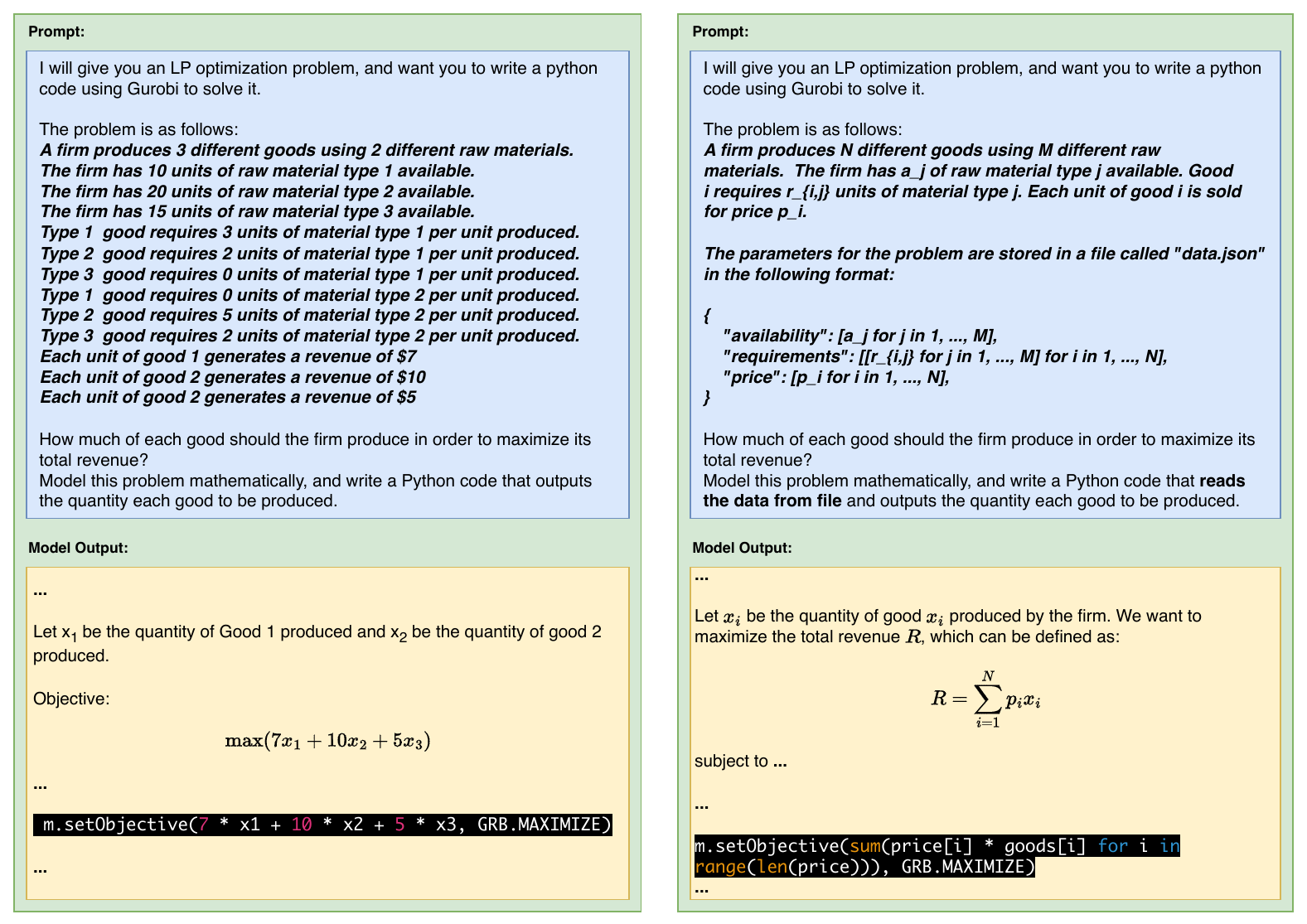}
    \caption{Scaling OptiMUS to problems with large numerical data: Instead of passing everything to the LLM directly (left), in OptiMUS we separate the numerical data from the problem description and give the metadata to the LLM (right). The LLM then writes a code to interact with the data file.}
    \label{fig:data_sep_example}
\end{figure}

\section{Methodology}
\label{methodology}

This section details the design of OptiMUS.
See \cref{fig:agent_structure} for an illustration.
OptiMUS starts with a structured description of the optimization problem, explained in \cref{snops}, and a separate data file.
It first transforms the structured description into 1) a mathematical formulation of the problem
and 2) tests that check the validity of a purported solution.
Afterwards, OptiMUS transforms the mathematical formulation into solver code. It joins the solver code with the problem data to solve the problem. If the code raises an error or fails a test, 
OptiMUS revises the code and repeats until the problem is solved or maximum iterations are reached. 
All prompts used by OptiMUS appear in \cref{prompts_appendix}. 

\subsection{Structured Natural language Optimization Problem (SNOP)}
\label{snops}

\begin{figure}[t]
    \centering
    \begin{subfigure}[b]{1\textwidth}
        \centering
        \includegraphics[scale=0.5]{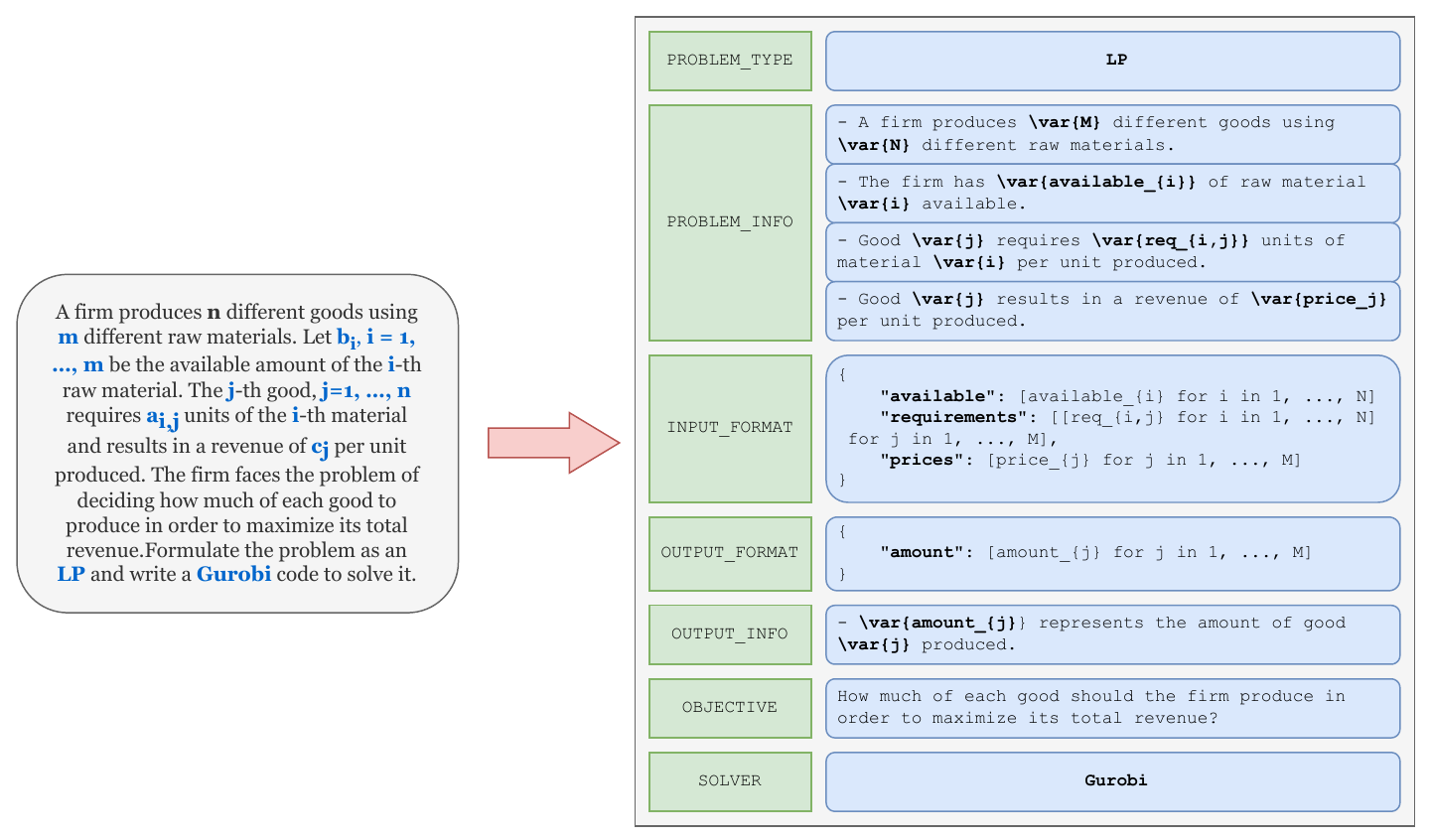}
        \caption{}
        \label{fig:text_to_snop}
    \end{subfigure}
    \begin{subfigure}[b]{0.45\textwidth}
        \centering
        \includegraphics[width=\textwidth]{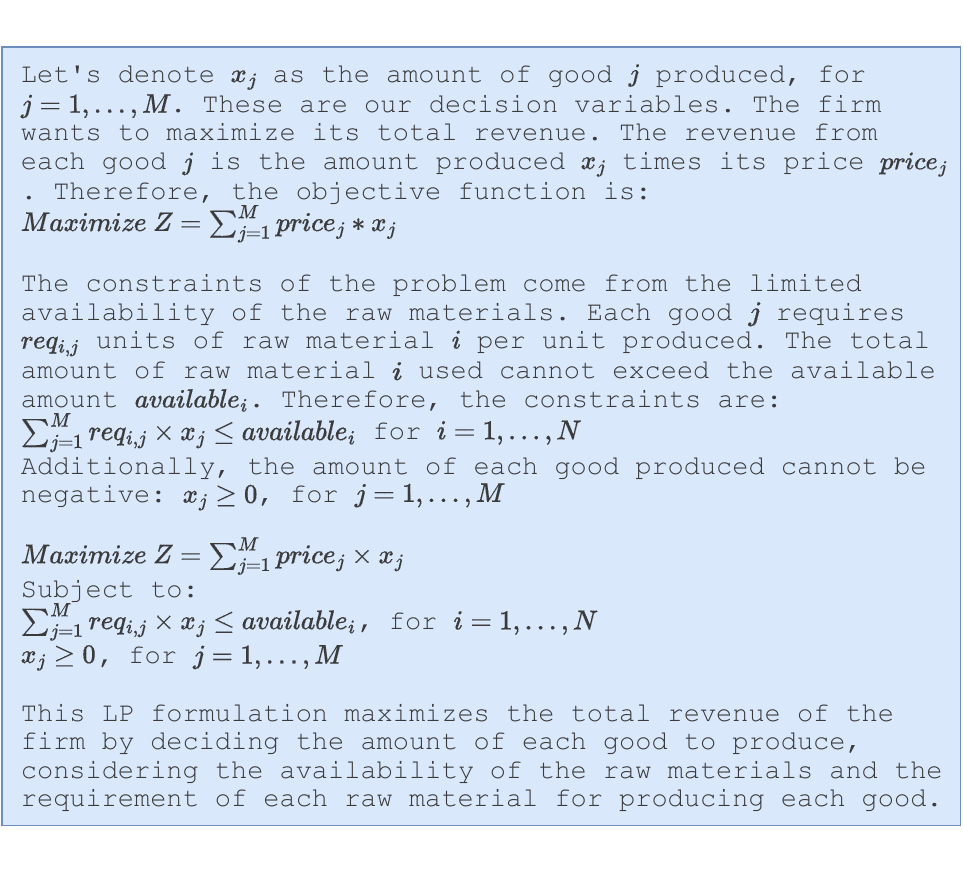}
        \caption{}
        \label{fig:latex_sample}
    \end{subfigure}
    \begin{subfigure}[b]{0.54\textwidth}
        \centering
        \includegraphics[width=\textwidth]{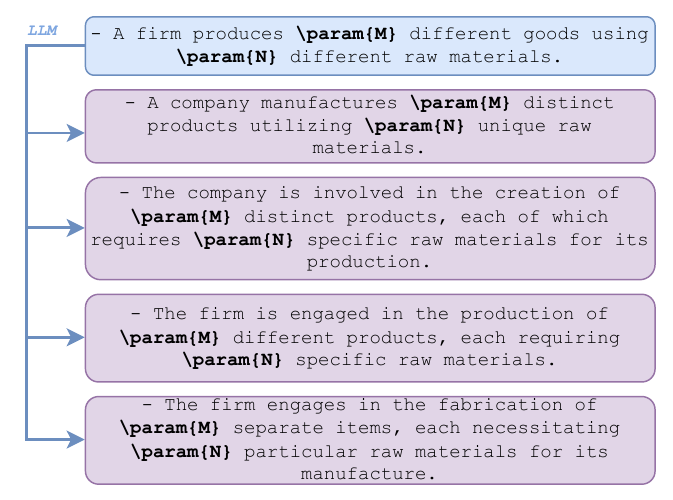}
        \caption{}
        \label{fig:aug_sample}
    \end{subfigure}
    
    \caption{a). An example of a real-world optimization problem and a SNOP representation for it. b). An example markdown { } formulation of a problem generated by OptiMUS. c) Example rephrasings generated by OptiMUS from a problem info statement in the augmentation process.}
    \label{fig:representation_format}
\end{figure}

As mentioned in \ref{challenges}, passing all the problem information directly to an LLM is not a scalable solution. To address this issue, we separate the data from the problem description, save the data as a \texttt{JSON} file, and then pass the format of this file to the LLM (an example is illustrated in \cref{fig:data_sep_example}). 
We use a standardized structure to organize our dataset instances, which we call Structured Natural language Optimization Problem (SNOP).
OptiMUS takes a SNOP as input, and our benchmark dataset contains SNOPs and their solutions.

A SNOP has 6 fields (see \cref{fig:representation_format}):

\begin{itemize}[leftmargin=*]
    \item \textbf{Problem type}: 
    The type of the problem (as a string), e.g., LP, MILP, QP, etc. 
    For example, by targeting LP rather than MILP, a user can instruct OptiMUS to relax an integer problem.
    This field can be set to \textit{ANY} to let OptiMUS decide how to formulate the problem.
    
    \item \textbf{Problem info}: A list of statements detailing the problem. We use \textit{\textbackslash param\{\}} symbol to identify problem parameters in the description. 
    The optimization solver will replace these by the problem data after the problem is formulated by the LLM.
    
    \item \textbf{Input format}: A string describing the format of the input data. We use \texttt{[]} to represent lists of values, names in quotes (\texttt{"}) to represent \texttt{JSON} keys, and \texttt{pseudo-for} to show indices.

    \item \textbf{Output info}: A list of statements describing the desired output.
    
    \item \textbf{Output format}: A string describing the format of the desired output.

    \item \textbf{Objective}: A string stating the objective of the optimization problem.

    \item \textbf{Solver}: The solver of choice (as a string), e.g., \texttt{Gurobi}, \texttt{cvxpy}, etc. 
    This field can be set to \textit{ANY} to let OptiMUS decide which solver to use.
\end{itemize}

Each problem in the benchmark has input data saved as a \texttt{\texttt{JSON}} file matching the input format. 
OptiMUS uses only the SNOP to develop a formulation, and then generates code to call an optimization solver, joining the formulation with the (possibly large) data to solve the problem. 

\subsection{Formulation}

Given a SNOP, OptiMUS generates a mathematical formulation for the problem in markdown. At this stage, OptiMUS prompts the language model to define the optimization variables, the constraints, and the objective function.

\subsection{Code Generation}

\begin{figure}
    \centering
    \includegraphics[width=\textwidth]{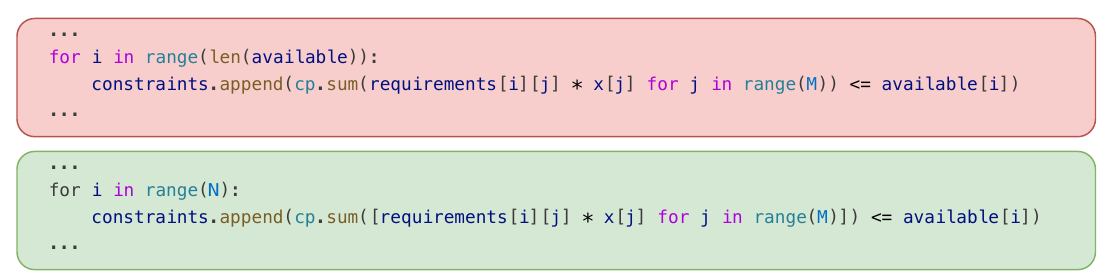}
    \caption{OptiMUS prompts include instructions to avoid common coding mistakes.
    For example, ChatGPT commonly uses \texttt{cvxpy.sum} on generator objects instead of lists. Adding the instruction "\textit{- cvxpy.sum takes a list as input, and not a generator}" to the code generation template reduces the incidence of this mistake. Top) generated code before the instruction; Bottom) generated code after adding the instruction.}
    \label{fig:solver-instructions}
\end{figure}

Given the problem formulation, OptiMUS generates \texttt{Python} code to read the input data from a {\texttt{JSON}} file, 
calls an optimization solver to solve the problem, 
and saves the output to another \texttt{\texttt{JSON}} file. 
OptiMUS uses \texttt{Gurobi} and \texttt{cvxpy} to solve the problems in our benchmark, and is capable of using advanced solver features. For example we observe it using \texttt{gurobi.abs\_} to model $\ell_1$-norm objective instead of adding auxiliary constraints and variables.
We observe that certain models make recurring mistakes when writing codes for certain solvers, and therefore our prompt includes solver-specific instructions to improve performance. For example, in \verb|cvpxy|, the model commonly uses \verb|cvxpy.sum| with generator objects instead of lists (see \cref{fig:solver-instructions}).

\subsection{Tests and revision}
\label{tests_and_revision}

Once solver code is generated, OptiMUS executes it with a \texttt{Python} interpreter.
Two outcomes are possible: 1) an execution error occurs or 2) an output is written to file. 
Given an execution error, the error string is passed to OptiMUS along with the code and formulation, so that OptiMUS can revise the solver code to fix the error. 
This process is repeated until the code successfully executes or maximum iterations are reached.

Given an output, we need to ensure its correctness.
OptiMUS generates unit tests using the SNOP to ensure the validity of the output. These tests check for 1) correct json formatting (e.g. the output json should contain ``amount'' as a key) 2) constraint satisfaction (e.g. detecting negative order quantities), and 3) consistency between the output values (e.g. sum of monthly profits should be equal to total profit)

\begin{figure}
    \centering
    \includegraphics[width=\textwidth]{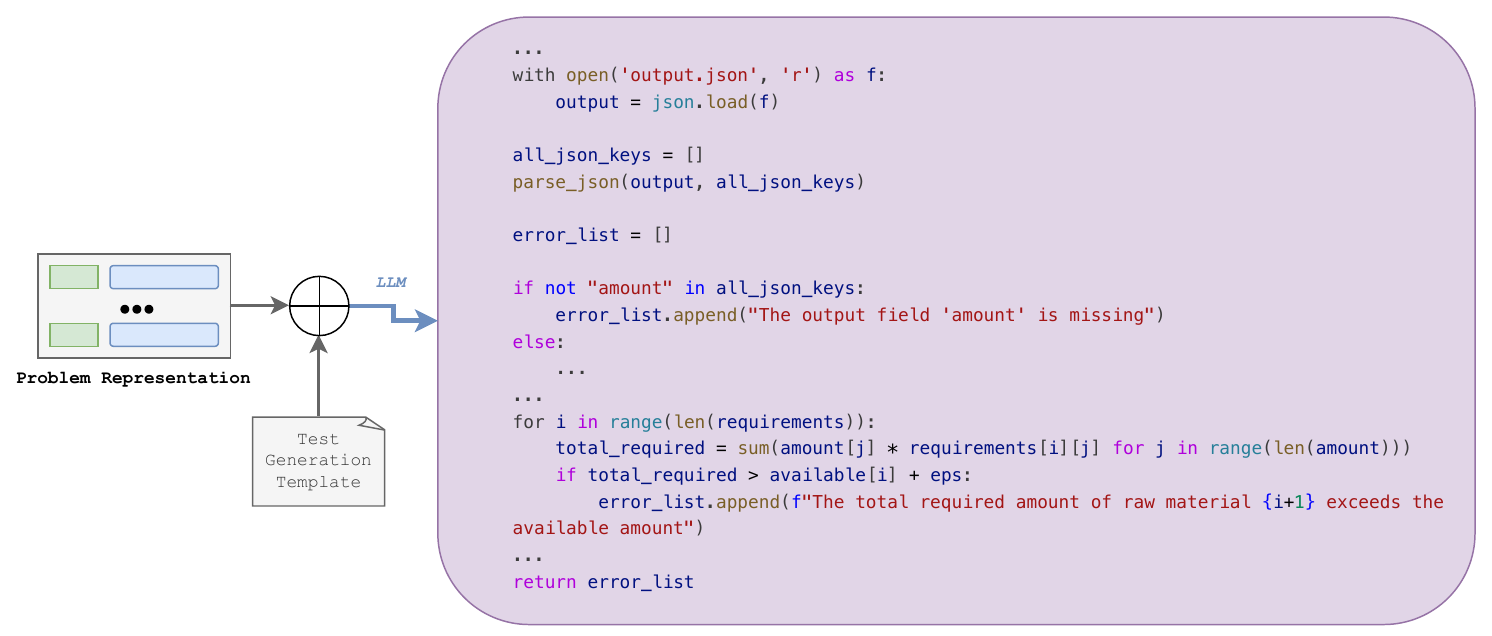}
    \caption{LLM can be used to generate tests and check the correctness of the output. After feeding the problem to an LLM using the test generation template, the model generates a script that checks the correctness of output (constraint satisfaction, output format, etc.) and returns appropriate error message if it finds a problem. The error message is used to automatically fix the code.}
    \label{fig:testgen-example}
\end{figure}

See \cref{fig:testgen-example} for an example.
Optionally, a user of OptiMUS can also revise these generated tests or write additional tests.
We call these auto-generated tests, supervised tests, and human tests, respectively.
Our benchmark includes supervised tests for every problem.
In our experience, developing supervised tests is roughly five times faster than developing equivalent human tests from scratch.
Given an output, OptiMUS runs the unit tests.
Any tests that fail will generate error messages. OptiMUS uses these error messages to revise the code and fix it. 

\subsection{Augmentation}
As an additional strategy, OptiMUS automatically rephrases problems and attempts to solve the rephrased version using the same workflow above. If any of the rephrased versions succeeds, OptiMUS is able to use the solution and hence solve the problem. See \cref{fig:aug_sample} for an example.

\section{Dataset}
\label{dataset}
To evaluate the performance of language models for solving optimization problems represented in natural language, we create a natural language optimization benchmark NLP4LP (Natural Language Processing for Linear Programming), consisting of 41 LP and 11 MILP problems (52 instances in total). The problems are drawn from textbooks and lecture notes in optimization \citep{intro_to_opt, model_building, lectures_in_lp_modeling}. 
These resources appeared before 2021, and there is a chance that parts of these books have been discussed on the internet and used to train LLMs. 
However, none of these textbooks include code snippets.
The natural language representations used in NLP4LP are further modified
from the original problem statement by representation as SNOPs
and by abstraction, 
as we replace numerical values in the original text by parameters.
Moreover, it is clear from our results that the LLMs still find it challenging to formulate and solve these problems.
The data consists of several views of each problem:
\begin{itemize}
\item SNOP (string) (see \cref{snops})
\item validity tests (code) (see \cref{tests_and_revision})
\item example data file (\texttt{JSON}) (see \cref{fig:data_output_example})
\item optimal value (floating point number)
\item sample optimal output (\texttt{JSON}) (see \cref{fig:data_output_example})
\end{itemize}

\section{Experiments and Analysis}

\label{experiments}

\begin{figure}[t]
    \centering
    \includegraphics[width=0.99\textwidth]{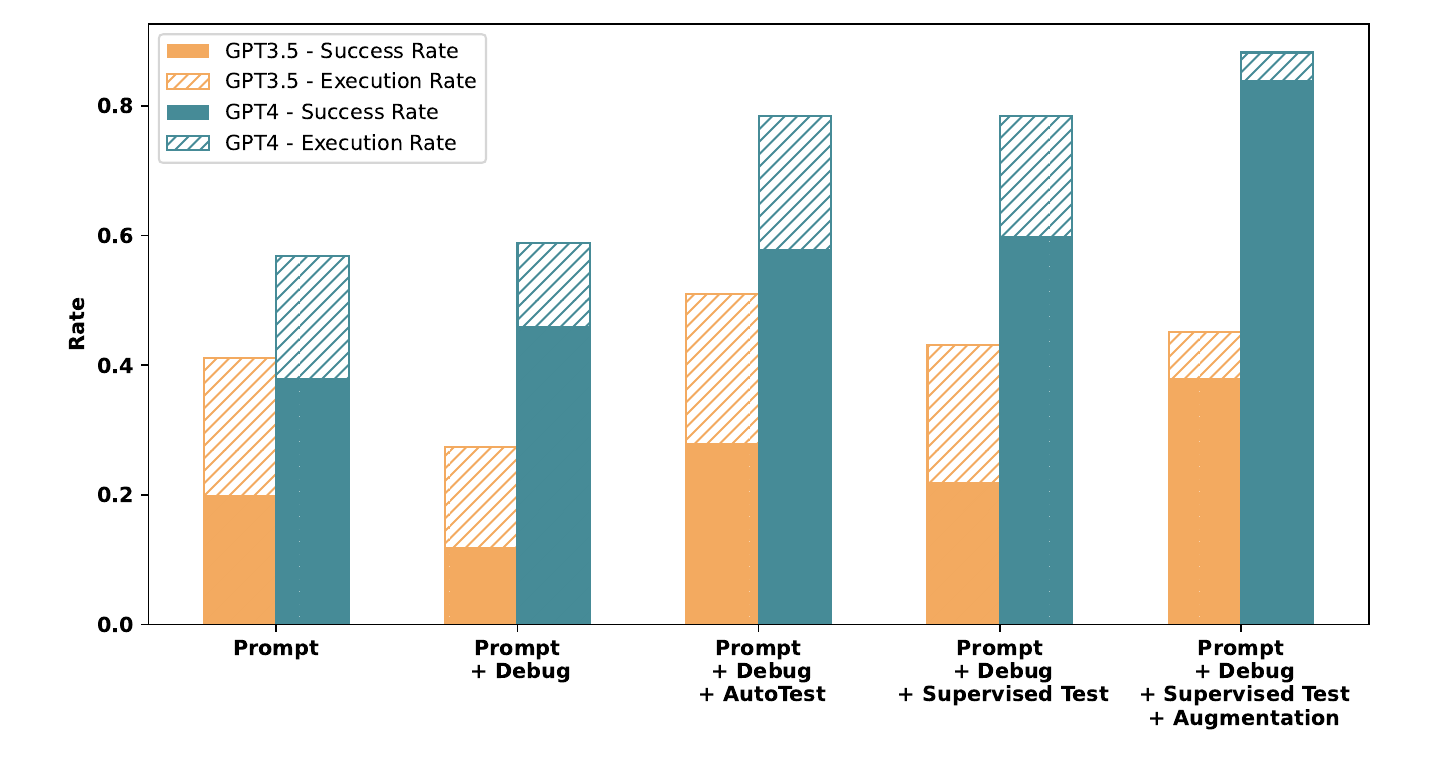}    
    \includegraphics[width=0.49\textwidth]{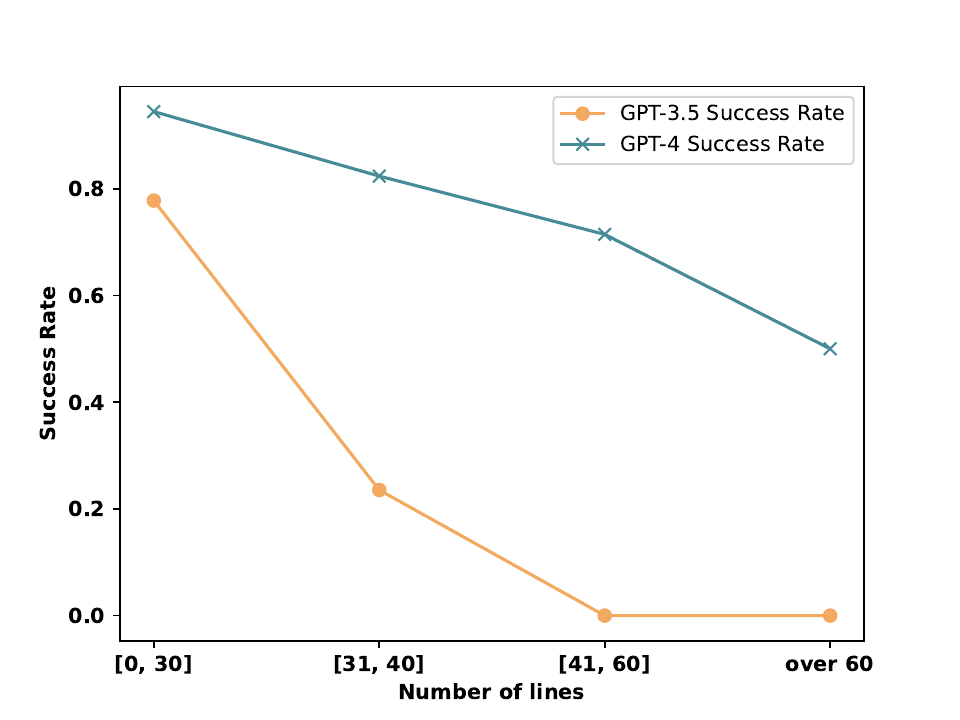}
    \includegraphics[width=0.49\textwidth]{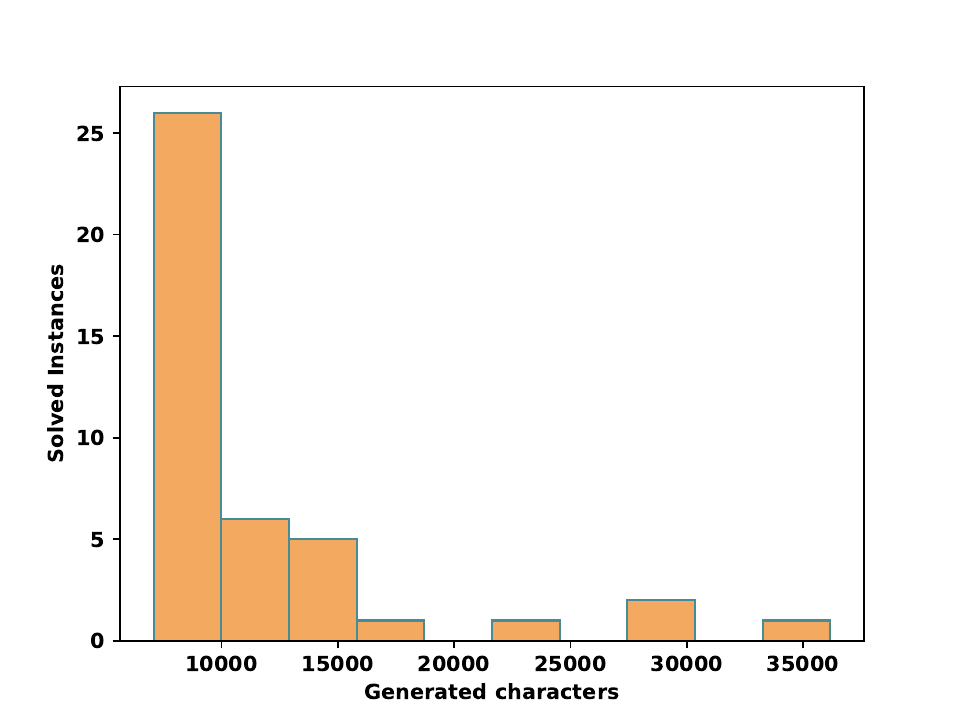}
    \caption{OptiMUS outperforms standard prompting by a large margin. Top: A comparison of the success rate and the execution rate for different modes on NLP4LP dataset using GPT-3.5 and GPT-4. Bottom left: Performance of OptiMUS vs the length of the SNOP. Bottom right: Distribution of the number of generated characters for solved instances using OptiMUS and GPT4}
    \label{fig:ablation}
\end{figure}

In this section, we empirically evaluate OptMus on NLP4LP and identify its strengths and weaknesses. We use GPT-3.5 and GPT-4 models for our experiments. The task of developing optimization formulations and solver code directly from natural language representations is new and there are no baselines in the literature. 
Therefore, we use simple prompting as a baseline and analyze the effect of adding each component. 
Concretely, we consider these five modes:

\begin{itemize}[leftmargin=*]
    \item Prompt: The problem is described in a few prompts and the LLM is asked to formulate the problem and write code to solve it using a given optimization solver. The code is run and the output is observed. We use this mode as a baseline.
    \item Prompt + Debug: In addition to the above, if the code raises syntax or runtime errors, the errors along with the code and the problem info are passed to the language model, which is prompted to debug the code. This cycle is repeated until the code runs or the maximum iterations are reached.
    \item Prompt + Debug + AutoTests: In addition to the above, when the code successfully runs and produces an output, automatically-generated tests are run to check the validity of the output. If the tests fail, the error messages are included with the problem description and passed back to the LLM for revision until the tests pass or the maximum iterations are reached.
    \item Prompt + Debug + Supervised Tests: Same as the above, except that the automatically-generated tests are all manually revised and fixed by experts if necessary.
    \item Prompt + Debug + Supervised Tests + Augmentation (OptiMUS): In addition to the above, each problem is rephrased using an LLM five times, and then the whole pipeline is applied to each of the rephrased versions independently. The problem is solved if at least one of the rephrased versions is solved.
\end{itemize}

We assess the models based on two metrics: success rate (the ratio of outputs satisfying all constraints and finding the optimal solution), and execution rate (the ratio of generated codes that are executable and generate an output). The results are available in \cref{fig:ablation}.

Using GPT-4, the success rate of our method improves with each of the additional features and capabilities described above. Basic prompting yields the lowest performance, as anticipated. Debugging improves the model’s ability to correct execution errors, while increasing the execution rate. Adding AutoTests improves the success rate  as it helps the model identify errors not found during the execution stage. Expert supervision to fix tests enhances performance but lowers the execution rate, as some revisions to fix test errors may inadvertently break the code. Finally, adding augmentations improves both success and execution rates, as OptiMUS can approach and solve problems in different ways, increasing the probability of finding a correct solution.

We observe a slightly different trend for GPT-3.5. Debugging improves the execution rate to some degree, but automated tests decrease performance due to the generation of incorrect, misleading tests. Compared to GPT-4, generating incorrect tests is  more common when using GPT-3.5 since it is a weaker model. Supervised test-fixing brings performance back to the level of simple prompting, showing that GPT-3.5 is not capable of correcting codes based on test error messages. However, augmentations significantly improve the success rate by giving the model multiple attempts to solve the problem with different rephrasings.

\begin{figure}[t]
    \centering
    \begin{subfigure}[b]{0.49\textwidth}
        \centering
        \includegraphics[width=1\textwidth]{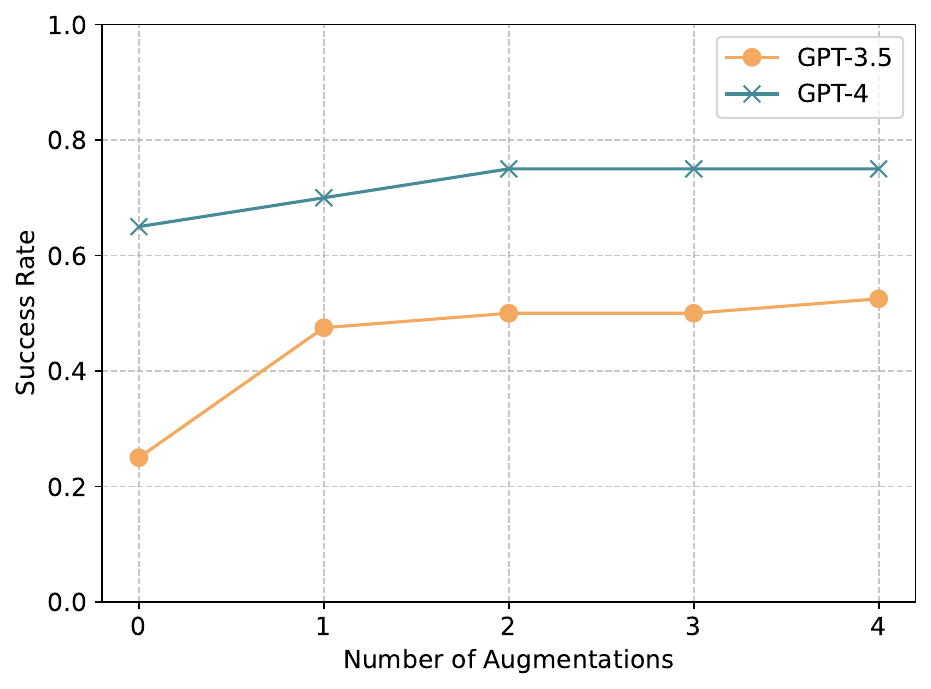}
        \caption{}
        \label{fig:aug_results}
    \end{subfigure}
    \begin{subfigure}[b]{0.49\textwidth}
        \centering
        \includegraphics[width=1\textwidth]{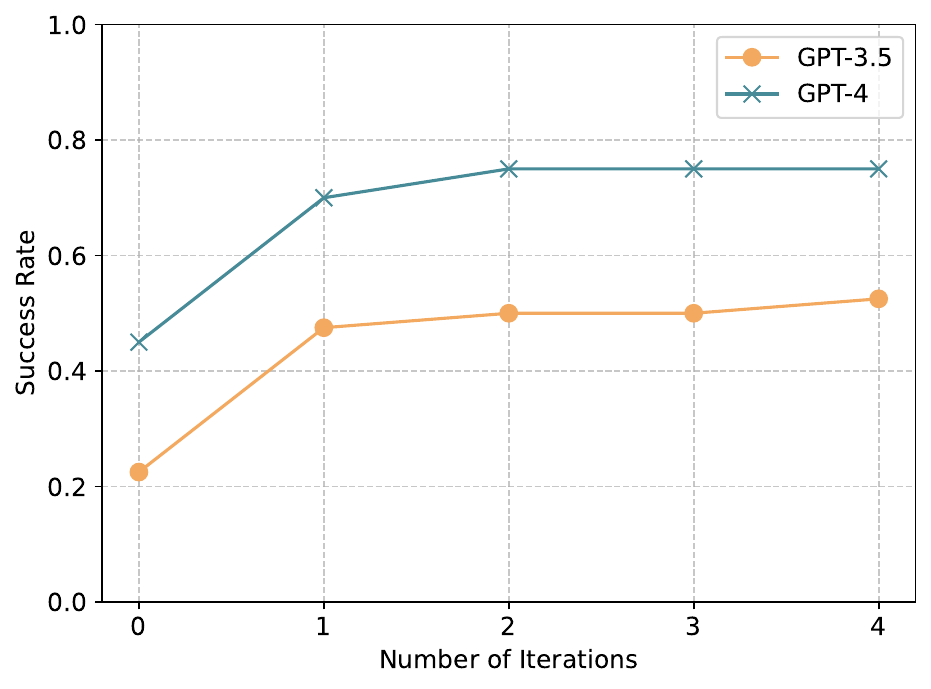}
        \caption{}
        \label{fig:iteration_results}
    \end{subfigure}
    \caption{Comparison of the success rate of OptiMUS with GPT-3.5 and GPT-4 for (a) different number of augmentations and (b) different number of iterations.}
    \label{fig:iterations_and_augmentations}
\end{figure}

More augmentations can improve the performance of OptiMUS at the cost of more computation. We observe that adding one or two augmentations increases the performance considerably, but additional augmentations beyond that point almost does not change OptiMUS's performance anymore (See \cref{fig:aug_results}). The reason is that further rephrasings after that point result in similar outputs to the initial ones. 

We observe a similar pattern for the maximum number of debugging iterations: one or two extra iterations can significantly improve performance, but increasing the maximum number of iterations beyond that point is not useful (See \cref{fig:iteration_results}). If the model is not able to solve the problem in its first few attempts, it usually gets stuck in an incorrect solution space and is often not able to find the right solution. Debugging step is capable of fixing minor errors like missing constraints or syntax/semantic errors, but it is not capable of fixing more fundamental errors like incorrect modeling problems.

Successful runs of the agent generate $4117.1 \pm 1509.7$ tokens on average. Given the fact that we used API calls to generate tokens, the formulation speed depends on factors like the internet speed, server responsiveness, model size, account priority, etc. In our experiments, all of the runs took less than 7 minutes (this is for all formulation and code/test generation steps, and does not include the solver run time). Note that OptiMUS can tackle augmented instances of the same problem in parallel.

\section{Related Work}
\label{related_work}
Many authors have considered the use of LLMs to solve mathematical problems. Most of the work in this domain is focused on training and tuning models on new datasets. \cite{frieder2023mathematicalChatGPTMath} introduce two datasets of math problems in algebra, counting, proof, calculus, probability, and various other topics to evaluate the performance of ChatGPT \citep{openai2023gpt4}. \cite{yuan2023wellArithmetic} propose an arithmetic dataset, MATH 401, to evaluate the performance of LLMs on arithmetic operations. \cite{minerva} further trains PaLM on a dataset of mathematical and scientific papers taken from arxiv. These papers aim to evaluate and improve the direct performance of LLM, but do not seek to integrate LLMs with other tools.

Other recent studies have explored ways to improve the accuracy and reach of LLMs by improving prompts and connecting LLMs to external tools. \cite{wei2023chainofthought} propose prompting the model to consider a series of intermediate reasoning steps to improve the performance of LLMs. They also use a calculator as an external tool to execute mathematical operations for LLMs. \cite{gao2022pal} uses an LLM to read natural language problems and generate programs as the intermediate reasoning steps, but offloads the solution step to a runtime such as a \texttt{Python} interpreter. \cite{he2023solvingMath} improve the performance of LLMs by delegating expression calculations to symbolic solvers. These methods take a step forward in augmenting LLMs with other tools.

However, these models are general-purpose and aim at a wide range of tasks. In contrast, our work exploits the particular structure of optimization problems to develop a new workflow for optimization modeling using LLMs together with optimization solvers that improves on simple prompting of LLMs.

Many research efforts aim to improve existing optimization solvers and develop new ones, 
using exact algorithms, heuristics, local and global search, simplex methods, branch and bound, dynamic programming, simulated annealing, and other methods \citep{adby2013introductiontoOptimization, koziel2011computationalOptMethods}. 
Deep learning has also been used to enhance optimization solvers \citep{BengioMLforComb, COwithGNNs, RlforCOSurvey}. 
Recent work has explored the possibility of using LLMs directly as solvers \citep{yang2023LLMsAsOptimizers}. 
In addition, ongoing work to enhance large language models (LLMs) \citep{openai2023gpt4, touvron2023llama, hoffmann2022training, vaswani2017attention} can support and enhance our results as the underlying LLMs improve.

\section{Conclusion}
\label{conclusion}

In summary, we developed OptiMUS, a Large Language Model (LLM)-based agent designed to formulate and solve optimization problems interpreted from natural language. We constructed NLP4LP, a novel dataset for optimization modeling from natural language, utilizing it to demonstrate the efficacy of the techniques implemented within OptiMUS. Our research serves as a proof of concept, illustrating the potential for automating various stages of the optimization procedure by leveraging LLMs together with traditional solvers.


Several avenues remain unexplored and can be further investigated. The quality of prompts can be enhanced through methods such as those proposed in \cite{yang2023LLMsAsOptimizers}. The NLP4LP dataset can be expanded by adding more instances and including other classes of optimization problems. It would be interesting to enable OptiMUS to work with unstructured natural language representations of problems instead of SNOPs. Moreover, performance of OptiMUS can be potentially improved by fine-tuning the LLM specifically for optimization problem modeling and solving.

\bibliography{iclr2024_conference}
\bibliographystyle{iclr2024_conference}

\newpage

\appendix

\section{Appendix}
\label{prompts_appendix}

\begin{figure}[h]
    \centering
    \includegraphics[width=0.98\textwidth]{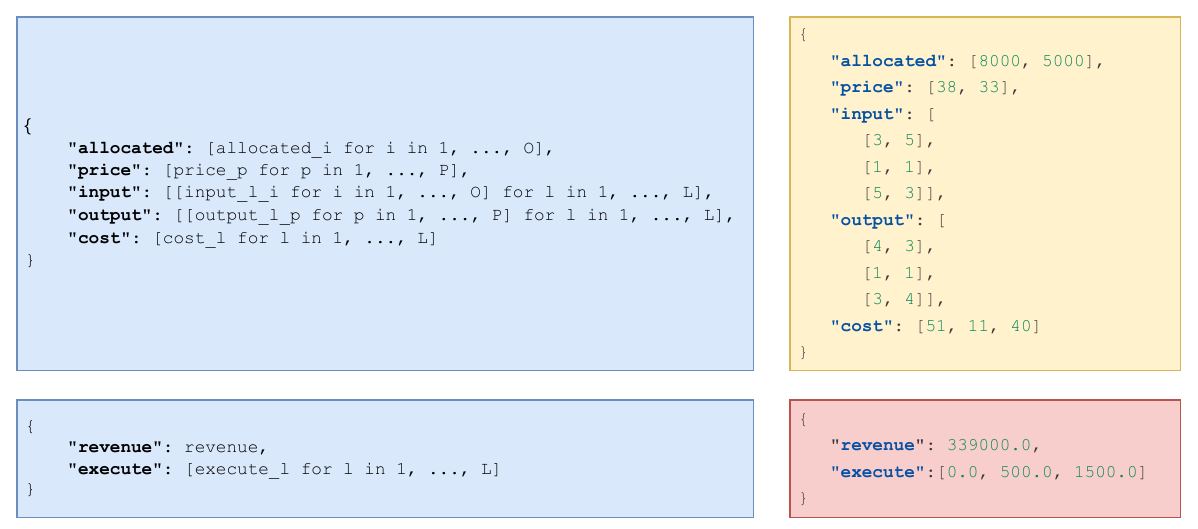}
    \caption{An example of input (top left) and output (bottom left) formats expressed in a SNOP and their equivalent data.json (top right) and output.json (bottom right) files.}
    \label{fig:data_output_example}
\end{figure}
\begin{figure}[t]
    \centering
    \includegraphics[width=\textwidth]{ 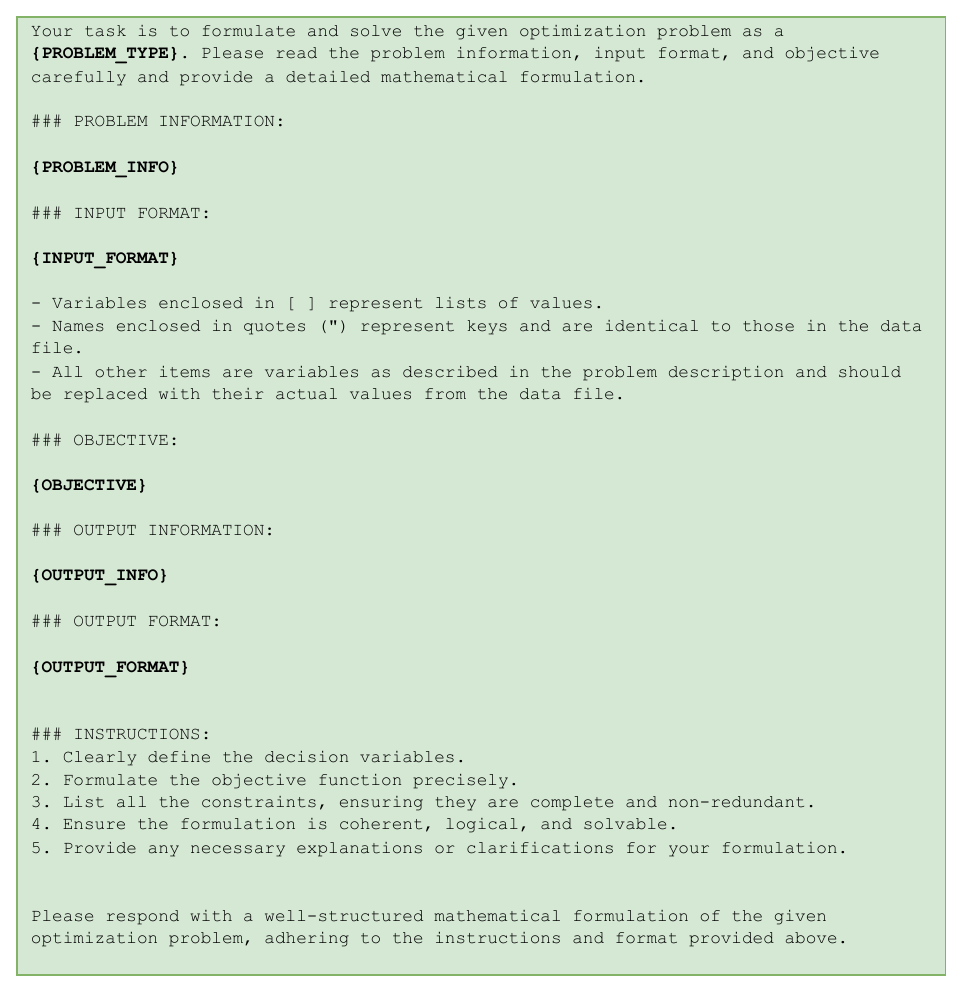}
    \caption{Template for formulating the optimization problems}
    \label{fig:formulation_template}
\end{figure}
\begin{figure}[t]
    \centering
    \includegraphics[width=\textwidth]{ 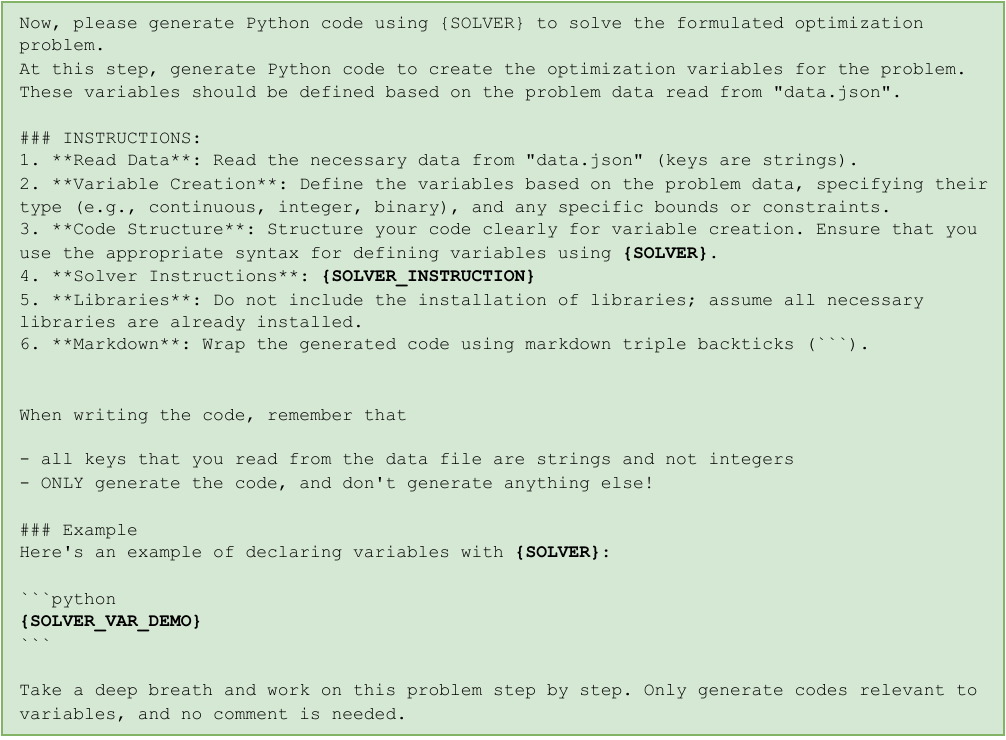}
    \caption{Template for generating code (variables) for the optimization problems}
    \label{fig:codegen_var}
\end{figure}
\begin{figure}[t]
    \centering
    \includegraphics[width=\textwidth]{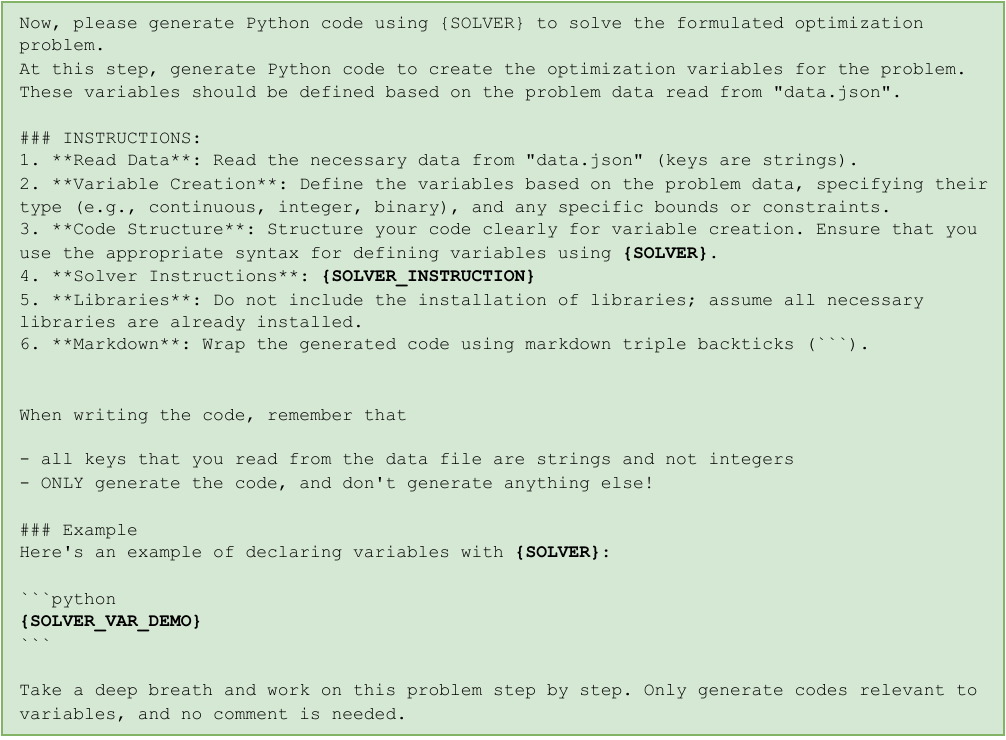}
    \caption{Template for generating code (constraints) for the optimization problems}
    \label{fig:codegen_constraint}
\end{figure}
\begin{figure}[t]
    \centering
    \includegraphics[width=\textwidth]{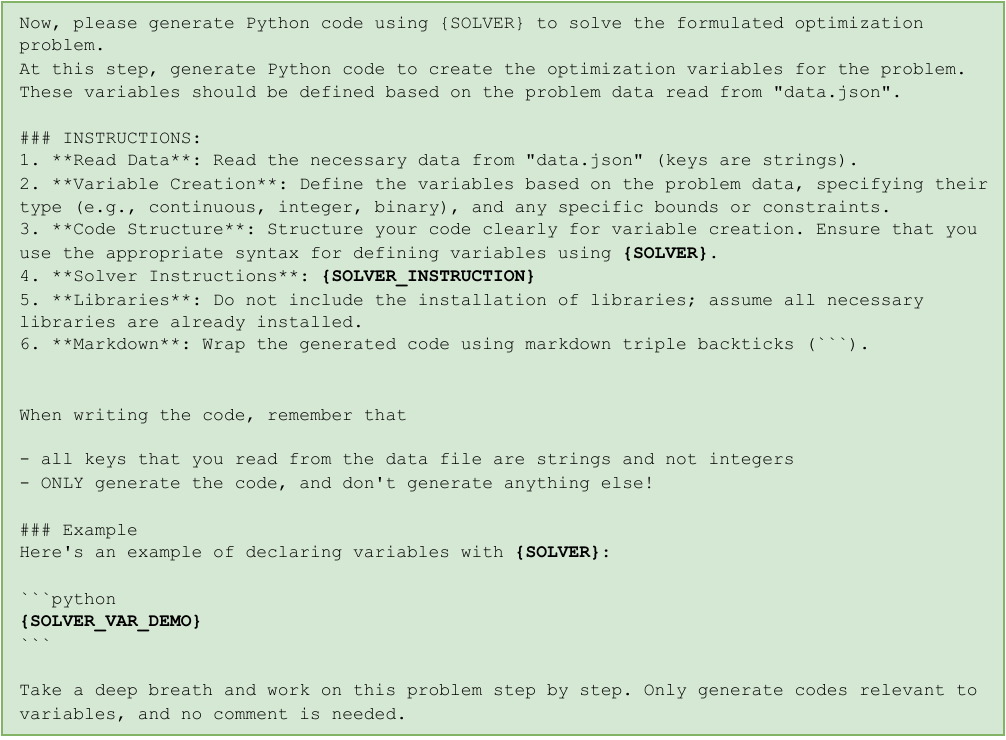}
    \caption{Template for generating code (objective) for the optimization problems}
    \label{fig:codegen_obj}
\end{figure}
\begin{figure}[t]
    \centering
    \includegraphics[width=\textwidth]{ 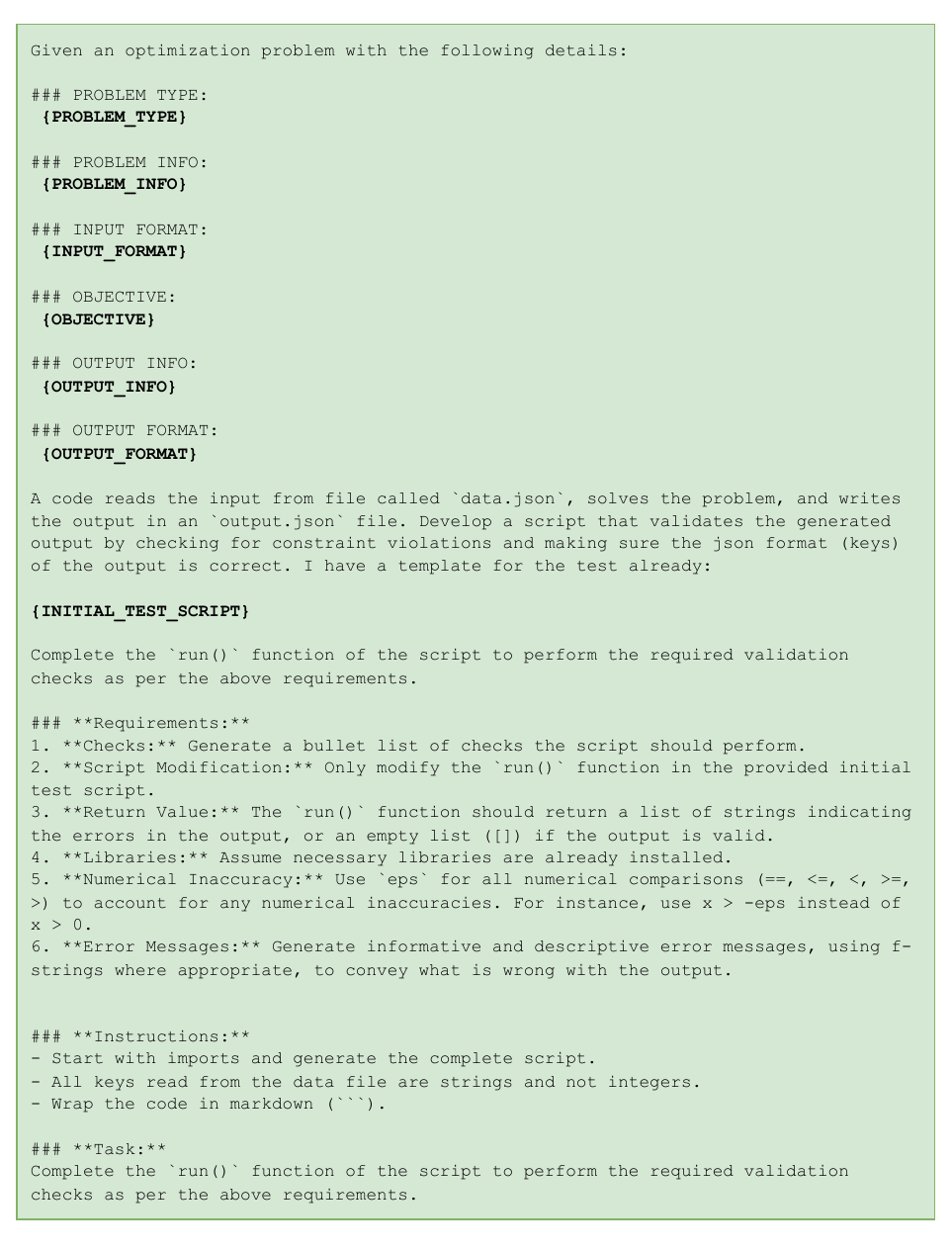}
    \caption{Template for generating tests for the optimization problems}
    \label{fig:testgen_tamplate}
\end{figure}
\begin{figure}[t]
    \centering
    \includegraphics[width=\textwidth]{ 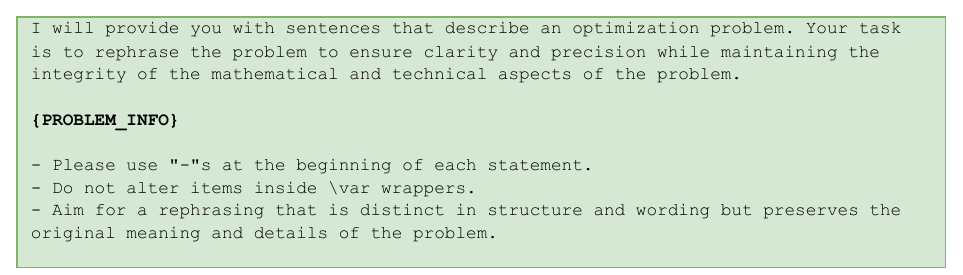}
    \caption{Template for rephrasing the optimization problems}
    \label{fig:rephrase_template}
\end{figure}
\begin{figure}[t]
    \centering
    \includegraphics[width=\textwidth]{ 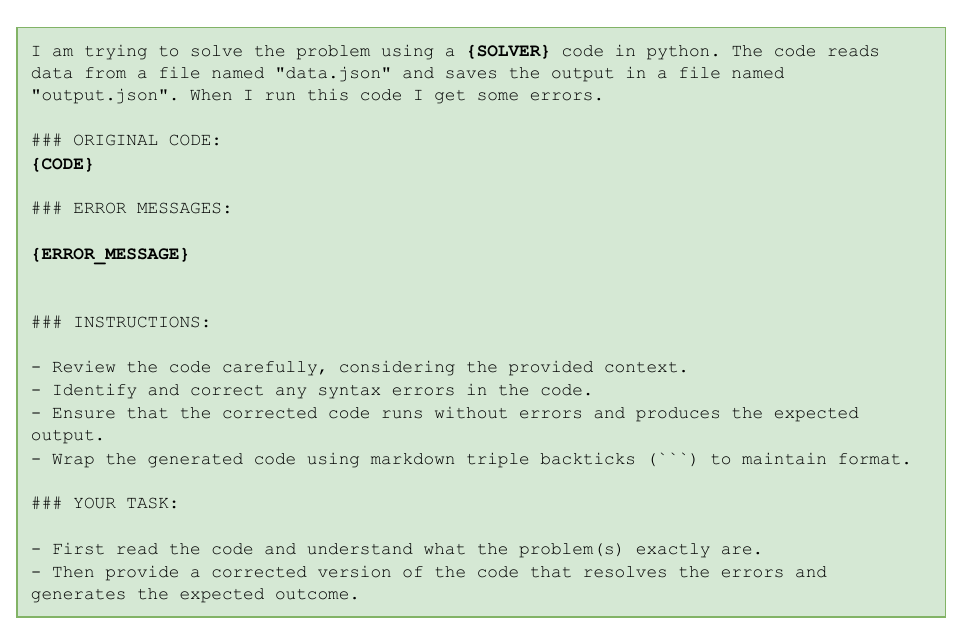}
    \caption{Template for debugging the code based on execution errors}
    \label{fig:debug_template}
\end{figure}
\begin{figure}[t]
    \centering
    \includegraphics[width=\textwidth]{ 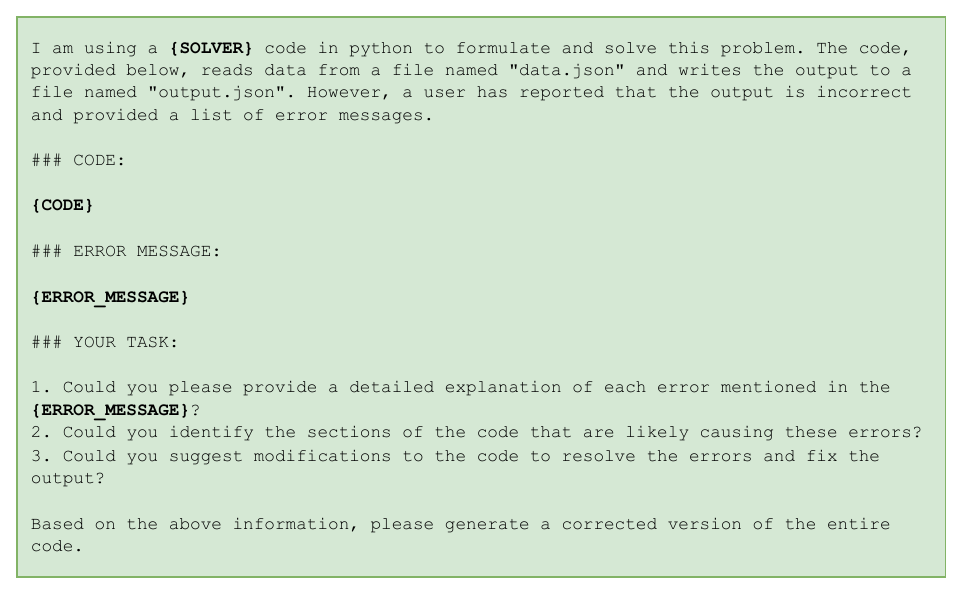}
    \caption{Template for fixing the code based on test errors}
    \label{fig:codefix_template}
\end{figure}

\end{document}